\documentclass[a4paper]{article}
\usepackage{eaclap}
\usepackage{epsfig}

\author{Caroline Lyon\\
Division of Computer Science\\ University of Hertfordshire\\
Hatfield AL10 9AB, UK\\
{\tt comrcml@herts.ac.uk}
\And
Bob Dickerson\\
Division of Computer Science\\ University of Hertfordshire\\
Hatfield AL10 9AB, UK\\
{\tt comqrgd@herts.ac.uk}}
\title{A fast partial parse of natural language sentences \\
using a connectionist method}
\begin{document}
\maketitle
\bibliographystyle{acl}
\raggedbottom
\begin{abstract}
The pattern matching capabilities of neural networks can be used
to locate
syntactic constituents of  natural language. This paper describes a fully
automated hybrid system, using neural nets operating within a grammatic
framework.
It addresses the representation of  language for
connectionist processing, and describes methods of constraining the problem
size. The function of the network is briefly explained, and results
are given.
\end{abstract}
\section{Introduction}
The pattern matching capabilities of neural networks can be used to detect
syntactic constituents of  natural language. This approach bears
comparison with probabilistic systems, but has the advantage that
negative as well as positive information can be modelled. Also, most
computation is done in advance, when the nets are
trained, so the run time computational load is low.
In this work neural networks are used as part of a fully automated system that
finds a
partial parse of declarative sentences. The connectionist processors operate
within a grammatic framework, and are supported by pre-processors that filter
the data and reduce the problem to a computationally tractable size.  A
prototype can be
accessed via
the Internet, on which users can try their own
text (details from the authors). It will take a sentence, locate
the subject
and then find the head of the subject. Typically 10 sentences take about
2 seconds, 50 sentences about 4 seconds, to process on a Sparc10 workstation.
Using the prototype on
technical manuals the subject and its head can be detected in over 90\%
of cases (See Section \ref{results}).

The well known complexity of parsing  is  addressed by
decomposing the problem, and then locating one syntactic constituent
at a time.
The sentence is first decomposed into the broad syntactic categories
\begin{center}
       pre-subject~~-~~subject~~-~~predicate
\end{center}
by locating the subject. Then these constituents can be processed further.
The underlying principle employed at each step is to take a sentence,
or part of a sentence, and generate strings with the  boundary markers of
the syntactic constituent in question placed in all possible positions. Then
a neural net selects the string with the correct placement.

This paper gives an overview of  how natural language is converted to a
representation that the neural nets can handle, and how the problem is
reduced to a manageable size. It then outlines the
neural net selection process. A comprehensive account is given in
\newcite{lyon1}; descriptions of the neural net process are also in
\newcite{lyon2} and \newcite{lyon3}.
This is a hybrid system. The core process is data driven, as the
parameters of the neural networks are derived from training text.
The neural net is trained in supervised mode
on examples that have been manually marked ``correct" and ``incorrect".
It will then be able to classify unseen examples.
However, the initial processing stages, in which the problem size is
constrained, operate within a skeletal grammatic framework.
Computational tractability is further addressed by
reducing data through the application of prohibitive rules as local
constraints.
The pruning process is remarkably effective.
\begin{figure*}[!bt]
\begin{center}
\epsfig{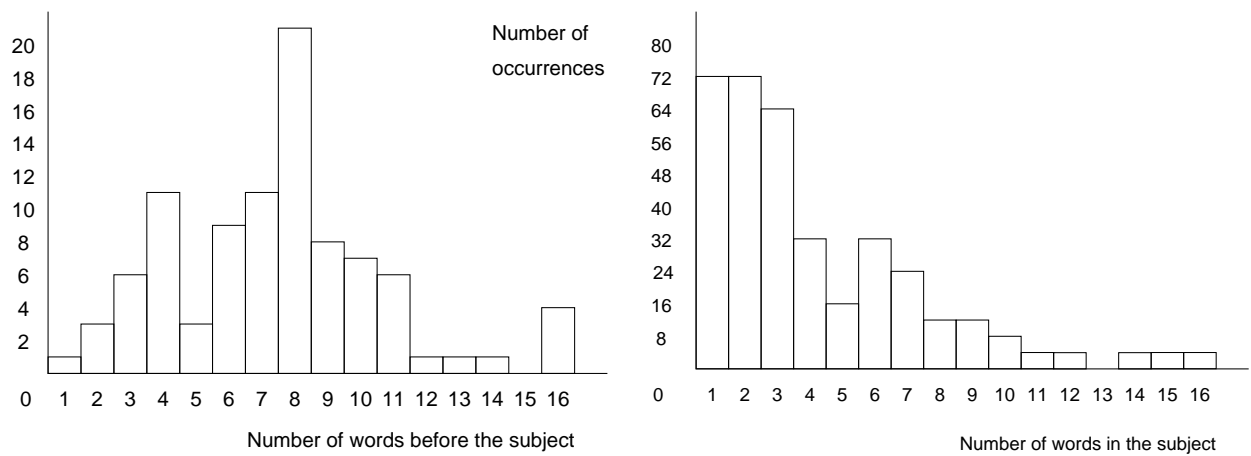}
\end{center}
\caption{\sf The frequency of constituent length for pre-subject and subject
in 351 sentences\label{histos}}
\end{figure*}

\section{The corpus of  sentences from technical manuals}
This work has principally been developed on text of technical manuals
from Perkins Engines Ltd., which have been translated by a semi-automatic
process~\cite{perkins}. Now, a partial parse can support such a process.
 For instance, frequently occurring modal verbs
such as ``must" are not
distinguished by number in English, but they are in many other languages.
It is necessary to locate the subject, then  identify the head and
determine its number in order to translate the main verb correctly in
sentences like (1) below.

\begin{quote}
\small
If a cooler is fitted to the gearbox, [ the pipe [ connections ] of the
cooler ] must be regularly checked for corrosion. ~~(1)
\end{quote}
\normalsize
\noindent
This parser has been trained to find the syntactic subject head that agrees in
number with the main verb.
The manuals are written using the PACE (Perkins Approved Clear English)
guidelines,
with the aim of producing clear, unambiguous texts. All declarative sentences
have been extracted for processing: about half were imperatives. This level of
classification can be done automatically in future. Table \ref{stats} and
Figure \ref{histos} show some of the characteristics of the corpus.
\begin{table}[h]
\begin{center}
\begin{tabular}{|l|c|}
\hline
Number of sentences & 351 \\ \hline
Average length & 17.98~words \\ \hline
No. of subordinate clauses & \\
In pre-subject & 65 \\
In subject & 19 \\
In predicate & 136 \\ \hline
Co-ordinated clauses & 50 \\ \hline
\end{tabular}
\end{center}
\small
Punctuation marks are counted as words, formulae as 1 word.
\normalsize
\caption{Corpus statistics\label{stats}}
\end{table}

\section{Language representation (I)}
\normalsize
In order to reconcile computational feasibility to empirical realism
an appropriate form of language representation is critical.
The first step in constraining the problem size is to partition an unlimited
vocabulary into a restricted number of part-of-speech tags.
Different stages of processing place different requirements on the
classification system, so customised tagsets have been developed. For the
first processing stage we need to place the subject markers, and,
as a further task, disambiguate tags. It was not found
necessary to use number information at this stage. For example,
consider the sentence:
\small
\begin{center}
\sf
Still waters run deep. ~~(2)
\end{center}
\normalsize
The word ``waters" could be a 3rd person, singular, present verb or a
plural noun. However, in order to
disambiguate the tag and place the subject markers it is only necessary to
know that it is a noun or else a verb.
The sentence parsed at the first level  returns:
\small
\begin{center}
\sf
[ Still waters ]  run deep. ~~(2.1)
\end{center}
\normalsize
The tagset used at this stage, mode 1, has 21 classes, not distinguished for
number. However, the head of the subject is then found
and number agreement with the verb can be assessed. At this stage the tagset,
mode 2, includes number information and has 28 classes.
Devising optimal tagsets for given tasks is a field in which further work is
planned. We need larger tagsets to capture more linguistic
information, but smaller ones to constrain the computational load.
Information theoretic tools can be used to find the entropy of
different tag sequence languages, and support decisions on
representation.

A functional approach is taken to tagging: words are
allocated to classes depending on their syntactic role. For instance,
superlative adjectives can act as nouns, so they are initially given the 2
tags: noun or adjective. This approach can be  extended by
taking adjacent words which act jointly as single lexical items
 as a unit. Thus the pair ``most~$<$adjective$>$" is taken as a
single superlative adjective.

Text is automatically tagged using the first modules of the CLAWS
program (1985 version), in which words are allocated one or more tags
from 134 classes \cite{garside}. These 134 tags are then mapped onto the
small customised tagsets. Tag
disambiguation is part of the parsing task, handled by the neural net and its
pre-processor. This version of CLAWS has a dictionary of about 6,300
words only. Other words are tagged using suffix information, or else defaults
are invoked. The correct tag is almost always included in the set allocated,
but more tags than necessary are often proposed.
A larger dictionary in  later versions  will address
this problem.

\subsection*{Representing syntactic boundary markers}
In the same way that tags are allocated to words, or to punctuation
marks, they can represent the boundaries of syntactic constituents, such as
noun phrases and verb phrases. Boundary markers can be considered
invisible tags, or hypertags, which have probabilistic
relationships with adjacent tags in the same way that words do.
\newcite{atwell4}
and \newcite{church}
 have used this approach.
If embedded syntactic constituents are
sought in a single pass, this can lead to computational overload
\cite{atwell6}.
Our approach uses a similar concept,
but differs in that embedded syntactic constituents are detected one at a
time in separate steps. There are only 2 hypertags - the
opening and closing brackets marking the possible location(s) of the
syntactic constituent in question.
Using this representation a hierarchical language structure
is converted to a string of tags represented by  a linear vector.
\begin{figure*}[t]
\begin{center}
\strut\psfig{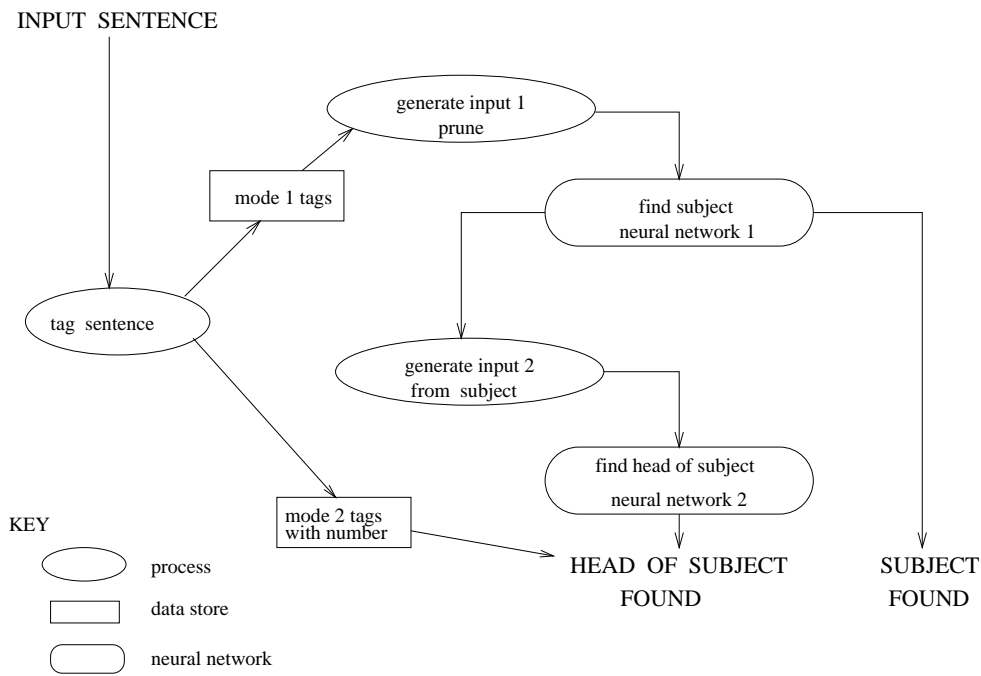}
\end{center}
\caption{Overview of the syntactic pattern recognition
process\label{fig1}}
\end{figure*}

\section{Constraining the generation of candidate strings}
\label{prune}
This system generates sets of tag strings for each sentence, with the
hypertags placed in all possible positions. Thus, for the subject
detection task:

\begin{quote}
\sf
\small
Then the performance of the pump must be monitored. ~~~~(3)
\end{quote}
\normalsize
\noindent
will generate strings of tags including:
\begin{quote}
\small
\sf
$[$ Then $]$ the performance of the pump must be monitored.~~ ~~($3.1$)\\
$[$ Then the $]$ performance of the pump must be monitored. ~~~~($3.2$)\\
.....................\\
Then $[$ the performance of the $]$ pump must be  monitored. ~~~~($3.n$)\\
Then $[$ the performance of the pump $]$ must be monitored. ~~~~($3.n+1$)\\
................
\end{quote}
\noindent
\normalsize
Hypertags are always inserted in pairs, so that closure is enforced.
There were arbitrary limits of a maximum of 10 words in the pre-subject and
10 words within the subject for the initial work described here.
These are now extended to 15 words in the pre-subject, 12 in the
subject - see Section \ref{results}. There must be at
least one word beyond the end of the subject and before the
end-of-sentence mark. Therefore, using the initial restrictions, in a sentence
of 22 words or more (counting punctuation marks  as words) there could be
100 alternative placements.
However, some words will have more than one possible tag. For instance, in
sentence~(1) above 5 words have 2 alternative tags, which will generate
$2^{5}$ possible strings before the hypertags are inserted. Since there are
22 words (including punctuation ) the total number of strings would be
\(2^{5}\ast 100 = 3200\).
It is not feasible to detect
one string out of this number: if the classifier marked {\em all} strings
incorrect the percentage wrongly classified  would only be $0.03\%$,
yet it would be quite useless. In order to find the correct string most
of the outside candidates must be dropped,
\subsection*{The skeletal grammatic framework}
A minimal grammar, set out in \newcite{lyon1} in EBNF form, is composed
of 9 rules. For instance, the subject must contain a noun-type word.
Applying this particular rule to sentence (3) above would eliminate
candidate strings (3.1) and (3.2). We also  have the 2 arbitrary
limits on length of pre-subject and subject.
There is  a small set of 4 extensions to the grammar, or
semi-local constraints. For instance, if a relative pronoun occurs, then
a verb must follow in that constituent.
On the technical manuals the constraints of the grammatic framework
put  up to 6\% of declarative sentences outside our system,
most commonly because the
pre-subject is too long. A small number are excluded because
the system cannot handle a co-ordinated head. With the  length of
pre-subject extended to 15 words, and subject to 12 words, an average
of 2\% are excluded (7 out of 351).

\subsection*{Prohibition tables}
The grammatic framework alone does not reduce the number of candidate
strings sufficiently for the subject detection stage. This problem
is addressed further by a method
suggested by \newcite{barton} that local constraints can
rein in the generation of an intractable number of possibilities.
In our system the local constraints are  prohibited tag pairs
and triples. These are adjacent tags which
are not allowed, such as ``determiner~-~verb" or ``start~of~subject~-~verb". If
during the generation of a candidate string a prohibited tuple is
encountered, then the process is aborted.
There are about 100 prohibited pairs and 120 triples.
By using these methods the number of candidate strings is
drastically reduced. For the technical manuals an average of 4 strings,
seldom more than 15 strings, are left. Around 25\% of sentences are left with
a single string.
 These filters or ``rules" differ fundamentally from generative
rules that produce allowable strings in a language. In those cases only
 productions that are explicitly admitted are allowed. Here, in
contrast, anything that is not expressly prohibited is allowed.
At this stage the data is ready to present to the neural net.
Figure \ref{fig1} gives an overview of the whole process.

\section{Language representation (II)}
Different network architectures have been investigated, but they all
share the same input and output representation. The output from the net
is a  vector whose 2 elements, or nodes, represent ``correct" and
``incorrect", ``yes" and ``no" - see Figure \ref{net-arch2}. The input
to the net is
derived from the candidate strings, the sequences of tags and
hypertags. These  must  be converted to  binary vectors. Each element of
the vector will represent a feature that is flagged 0 or 1, absent or
present.

Though the form in which the vector is written may give an
illusion of representing order, no sequential order is maintained.
A method of representing a sequence must be chosen. The sequential order of the
input is captured here, partially,
by taking adjacent tags, pairs and triples, as the feature elements. The
individual tags are converted to a bipos and tripos representation.
Using this method each tag is in 3 tripos and 2 bipos elements. This
highly redundant code will aid the processing of sparse data typical
of natural language.

For most of the work described here the sentence was dynamically truncated
2 words beyond the hypertag marking the close of the subject.
This process has now been improved by going
further along the sentence.
\begin{figure*}[t]
\begin{center}
\strut\epsfig{figure=net2.eps,width=13cm}
\caption{The single layer net: showing the feed forward process
\label{net-arch2} }
\end{center}
\end{figure*}

\section{The function of the net}
\label{net}
The net that gave best results was a simple single layer net
(Figure \ref{net-arch2}), derived from the Hodyne net of
\newcite{wyard}. This is conventionally  a
``single layer" net, since there is one layer of processing nodes.
Multi-layer networks, which can process linearly inseparable data,
 were also investigated, but are not necessary for
this particular processing task. The linear separability of data is related
to its order, and this system uses higher order pairs and triples as input. The
question of appropriate network architecture is examined in
\newcite{pao2}, \newcite{widrow1} and \newcite{lyon1}.

\subsection*{The training process}
The net is presented with training strings whose desired
classification has been manually marked. The weights on the connections
between input and output nodes are adjusted until a required level of
performance is reached. Then the weights are fixed and the trained net
is ready to classify
unseen sentences. The prototype accessible via the Internet has been trained on
sentences from the technical manuals, slightly augmented.

Initially the weighted links are disabled.
When a string is presented to the network in training mode, it
activates a set of input nodes. If an input node is not already linked
 to the output node representing the desired response, it will be connected
and the weight on the connection will be initialised to 1.0.
Most input nodes are connected to both outputs, since most tuples occur in
both grammatical and ungrammatical strings. However, some will only be
connected to one output - see Figure \ref{net-arch2}.

The input layer potentially has a node for
each possible tuple. With 28 tags, 2 hypertags and a start symbol
 the upper bound  on the number of input nodes is $31^3 + 31^2$.
In practice the maximum activated is currently about 1000. In testing mode,
if a previously unseen tuple appears it makes zero contribution to the result.
The activations at the input layer are fed forward through the weighted
connections to the output nodes, where they are summed. The highest  output
marks the winning node.  If the desired node wins, then no action is
taken. If the desired node does not win, then the weight on connections
to the desired node are incremented, while the weights on connections to
the unwanted node are decremented.

This algorithm differs from some commonly used methods. In feed forward
networks trained in supervised mode to perform a classification task
different penalty measures can be used to trigger a weight update. Back
propagation and some single layer training methods typically minimise a metric
based on
the least squared error (LSE) between desired and actual activation of
the output nodes. The reason why a differentiable error measure of this
sort is necessary for multi-layer nets is well documented, for example
 see \newcite{pdp}.  However, for single layer nets we can
choose to update weights directly: the error at an output node can trigger
weight updates on the connections that feed it. Solutions with LSE are
not necessarily the same as minimising the number of misclassifications,
and for certain types of data this second method of direct training may
be appropriate. Now, in the natural language domain it is desirable to
get information from infrequent as well as common events. Rare events,
rather than being noise, can make a useful contribution to a
classification task. We need a method that captures information from
infrequent events, and adopt a direct measure of misclassification.
This may be better suited to data with a ``Zipfian'' distribution
\cite{shannon}.

The update factor is chosen to meet several requirements.
It should always be positive, and asymptotic to
maximum and minimum bounds. The factor should be
greatest in the central region, least as it
moves away in either direction. We are currently
still using the original Hodyne function because
it works well in practice.
The update factor is given in the following formula.
If $\delta $= +1 for strengthening weights and $\delta$ = -1 for
weakening them, then
\[ w_{new} = \left[ 1 +
\frac{\delta \ast w_{old}}{1 + ( \delta \ast w_{old})^4} \right] w_{old}
 \]
Recall that weights are initialised to 1.0. After training we find that the
weight range is bounded by
\[10^{-3} < w < 5.0 \]
Total time for training is measured in seconds.
The number of iterative cycles that
are necessary depends on the threshold chosen for the trained net to
cross, and on details of the vector representation. The demonstration
prototype takes about 15 seconds. With the most recent
improved representation  about 1000
strings can be trained in 1 second, to 97\%. The results from using these nets
are given in Table \ref{final}. It was found that triples alone gave as good
results as pairs and triples together. And though the nets easily train to 99\%
correct, the lower threshold gives slightly better generalisation and
thus gives better results on the test data.
\subsection*{The testing process}
When the trained net is run on unseen data the weights on the links
are fixed. Any link that is still disabled is activated and
initialised to 0, so that tuples
which have not occurred in the training corpus make no contribution to
the classification task. Sentences are put through the
pre-processer one at a time and the candidate strings which are generated
are then presented to the network.
The output is now interpreted
differently.
The difference between the ``yes" and ``no" activation levels
is recorded for each string, and this score is considered a measure of
grammaticality, $\Gamma$.
The string with the highest $\Gamma$ score is taken as the correct
one.

For the results given below, the networks were trained on part of the corpus
and tested on another part of the corpus. For the prototype in which users
can process their own text, the net was trained on the whole corpus, slightly
augmented.
\begin{table*}[t]
\begin{center}
\begin{tabular}{|c|c||c|c|c|}
\hline
no. of          & no. of    &\% sents with  &\% sents   &\% sents with\\
training sents. &test sents. & subject     & correct        &subject and head\\
                 &           & found       & measure A          & found \\
\hline
220 &    42  & 100 & 100  &  95     \\\hline
 198 &    63  & 97  & 97   &  90    \\ \hline
 204 &    58  & 95 & 95   &  93    \\ \hline
 276 &    50  & 94  &    & 94    \\ \hline
\end{tabular}
\caption{Performance on text from Perkins manuals,  after 6\% sentences have
been excluded \label{all}}
\end{center}

\begin{center}
\begin{tabular}{|c|c||c|c|c|}
\hline
no. of          & no. of & \% sents with &\% sents  &\% sents with\\
training sents. &test sents.& subject   & correct      &subject and head\\
                 &           & found  &  measure A      &found        \\
\hline
 309 &    42  & 100 & 97.6  & 97.6   \\ \hline
 288 &    63  & 98.4 & 96.8   &96.8    \\ \hline
 292 &    59  & 98.3 & 98.3  & 96.6     \\\hline
 284 &    67  & 94.0 & 94.0   &  94.0    \\ \hline
\end{tabular}
\caption{Performance on text from Perkins manuals, using improved
representation and larger training set, after 2\% sentences have been
excluded\label{final}}
\end{center}
\end{table*}

\section{Results}
\label{results}
There are several measures of correctness
that can be taken when results are evaluated. The most lenient
is  whether or not the
subject and head markers are placed correctly - the type of measure used
in the IBM/Lancaster work \cite{ibm}. Since we are working
towards a hierarchical language structure, we may want the
words within constituents correctly tagged, ready for the next stage of
processing. ``correct- A" also requires that the words within the subject
are correctly tagged. The results in Tables \ref{all} and \ref{final}
give an indication of
performance levels.
 \section{Using negative information}
When parses are postulated for a sentence negative as well as positive
examples are
likely to occur. Now, in natural language   negative
correlations  are an important source of
information: the occurrence of some words or groups of words inhibit
others from following. We wish to exploit these
constraints.
\newcite{brill} recognised this, and introduced  the idea of
{\em distituents}. These are elements of a sentence that should be separated,
as opposed to elements of
{\em constituents} that cling together.
Brill addresses the problem of finding a valid metric for distituency
by using a generalized mutual information statistic.
 Distituency is marked by a mutual information minima. His method is
supported by a small 4 rule grammar.

However, this approach does not
fully capture the sense in which inhibitory factors play a negative and not
just a neutral role. We want to distinguish between items that are unlikely
to occur ever, and those that have just not happened to turn up in the
 training data. For example, in sentence (3) above
strings 3.1, 3.2 and 3.n  can never
be correct.  These should be distinguished from possibly correct parses that
are not in the training data.
In order that  ``improbabilities" can be modelled by inhibitory connections
\newcite{niles} show how a  Hidden Markov Model can be implemented by a
neural network.

The theoretical ground for incorporating negative examples in a
language learning process originates with the work of
\newcite{gold1}, developed by \newcite{angluin2}.
He examined the process of learning the grammar of a formal language
from examples. He showed that, for languages at least as high in the
Chomsky hierarchy as CFGs, inference from positive data alone is strictly less
powerful than inference from both positive and negative data together.
To illustrate this informally
consider a case of inference from a number of
examples: as they are presented to the inference machine, possible
grammars are postulated. However, with positive data alone a problem of
over generalization
arises: the postulated grammar may be a superset of the real grammar, and
sentences that are outside the real grammar could be accepted. If both
positive and negative data is used, counter
examples will reduce the postulated grammar so that it is nearer the
real grammar.
Gold developed his theory for formal languages: it is argued that similar
considerations apply here.
A grammar may be inferred from positive examples alone for certain subsets
of regular languages \cite{garcia}, or an inference process may
degenerate into a look up procedure
if every possible positive example is stored. In these
cases negative information is not required, but they are not plausible models
for unbounded natural language.
In our method the required parse is found by inferring the grammar
from both positive and negative information, which is effectively modelled
by the neural net.
Future work will investigate the effect of training the networks on the
positive examples alone. With our current size corpus there is not enough data.
\subsection*{Relationship between the neural net and prohibition table}
The relationship between the neural net and the
rules in the prohibition table should be seen in the following way. Any
single rule prohibiting a tuple of adjacent tags could be omitted and the
neural network
would handle it by linking the node representing that tuple to ``no"
only. However, for some processing steps we
need to reduce the number of candidate tag strings presented to the
neural network to  manageable proportions (see Section \ref{prune}).
The data must be pre-processed by
filtering through the prohibition rule constraints. If the number
of candidate strings is within desirable bounds, such as for the head
detection task, no rules are used. Our system is  data driven as far as
possible: the rules are invoked if they are needed to make the problem
computationally tractable.
\section{Conclusion}
Our working prototype indicates that the methods described here are
worth developing, and that connectionist methods can be used to generalise from
the training corpus to unseen text.
Since data can be represented as higher order tuples, single layer networks
can be used. The traditional problems of training times do not arise. We have
also used multi-layer nets on this data: they have no advantages, and perform
slightly less well \cite{lyon1}.

The supporting role of the grammatic framework and the prohibition filters
should not be underestimated. Whenever the scope of the system is extended
it has been found necessary to enhance these elements.

The most laborious part of this work is preparing the training data.
Each time the
representation is modified a new set of strings is generated that need
marking up. An autodidactic check is now included which speeds up this task.
We run marked up training data through an early version of the network
trained on the same data, so the results should be almost all correct.
If an ``incorrect'' parse occurs we can then check whether that sentence was
properly marked up.

Some of the features of the system described here could be used in
a stochastic process. However, connectionist methods have
 low computational loads at runtime. Moreover, they can utilise more of
the implicit information in the training data by modelling negative
relationships.
This is a powerful concept that can be
exploited in the effort to squeeze out every available piece of useful
information for natural language processing.

Future work is planned to extend this very limited partial parser, and
decompose
sentences further into their hierarchical constituent parts. In order to do
this a number of subsidiary tasks will be addressed. The system
is being improved by identifying groups of words that act as single
lexical items. The decomposition of the
problem can be investigated further: for instance, should the tag
disambiguation
task precede the placement of the subject boundary markers in a separate
step? More detailed
investigation of language representation issues will be undertaken. And
the critical issues of investigating the most appropriate network architectures
will be carried on.

\end{document}